\documentclass[letterpaper]{article} 
\usepackage{aaai25}  
\usepackage{times}  
\usepackage{helvet}  
\usepackage{courier}  
\usepackage[hyphens]{url}  
\usepackage{graphicx} 
\usepackage{natbib}  
\usepackage{caption} 
\usepackage{algorithm}
\usepackage{algorithmic}

\usepackage{newfloat}
\usepackage{listings}
\DeclareCaptionStyle{ruled}{labelfont=normalfont,labelsep=colon,strut=off} 
\floatstyle{ruled}
\newfloat{listing}{tb}{lst}{}
\floatname{listing}{Listing}
\usepackage{booktabs}
\usepackage{multirow}
\usepackage{tabularx}
\newcolumntype{Y}{>{\centering\arraybackslash}X}
\usepackage{colortbl}
\definecolor{LightCyan}{rgb}{0.88,1,1}
\usepackage[accsupp]{axessibility}
\usepackage{bbold}
\usepackage{siunitx}
\usepackage{subcaption}

\AtEndPreamble{
    \usepackage[capitalize]{cleveref}
    \crefname{section}{Sec.}{Secs.}
    \Crefname{section}{Section}{Sections}
    \crefname{table}{Tab.}{Tabs.}
    \Crefname{table}{Table}{Tables}
}

\title{Difficulty-aware Balancing Margin Loss for Long-tailed Recognition}
\author{
    Minseok Son\equalcontrib,
    Inyong Koo\equalcontrib,
    Jinyoung Park,
    Changick Kim
}
\affiliations{
    Korea Advanced Institute of Science and Technology\\

    \{ksos104,
    iykoo010,
    jinyoungpark,
    changick\}@kaist.ac.kr
}

\usepackage{bibentry}

\begin{document}

\maketitle

\begin{abstract}
When trained with severely imbalanced data, deep neural networks often struggle to accurately recognize classes with only a few samples.
Previous studies in long-tailed recognition have attempted to rebalance biased learning using known sample distributions, primarily addressing different classification difficulties at the class level.
However, these approaches often overlook the instance difficulty variation within each class.
In this paper, we propose a difficulty-aware balancing margin (DBM) loss, which considers both class imbalance and instance difficulty.
DBM loss comprises two components: a class-wise margin to mitigate learning bias caused by imbalanced class frequencies, and an instance-wise margin assigned to hard positive samples based on their individual difficulty.
DBM loss improves class discriminativity by assigning larger margins to more difficult samples.
Our method seamlessly combines with existing approaches and consistently improves performance across various long-tailed recognition benchmarks.
\end{abstract}

\begin{links}
    \link{Code}{https://github.com/quotation2520/dbm_ltr}
\end{links}

\section{Introduction}
\label{sec:intro}

In recent decades, deep neural networks have demonstrated remarkable success in image recognition tasks \cite{vgg_simonyan, resnet_he, googlenet_Szegedy}, largely due to the availability of large-scale datasets like ImageNet \cite{imagenet_deng}.
However, real-world datasets often exhibit an imbalanced distribution, known as a long-tailed distribution, wherein a few `head' classes contain a large number of samples, while numerous other classes, referred to as `tail' classes, contain significantly fewer samples.
This imbalance presents significant challenges: deep learning models, predominantly trained on the abundant majority classes, struggle to effectively learn features for the minority classes.
As a result, models tend to underperform on these underrepresented classes, compromising their overall accuracy.

\begin{figure}[!t]
    \centering
    \includegraphics[width=\columnwidth]{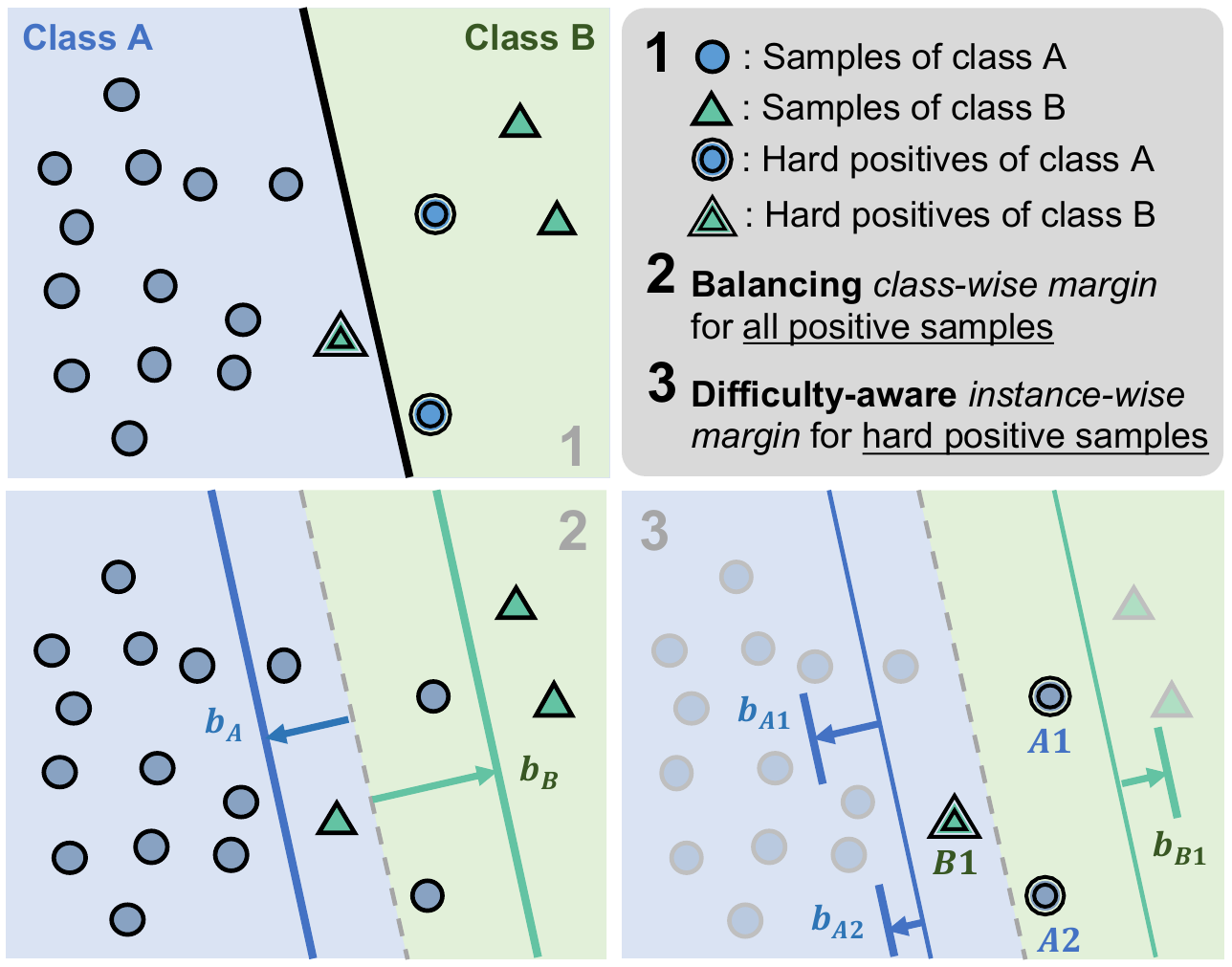}
    \caption{
    Overview of our method. 
    The model is trained to align samples within decision boundaries defined by adaptive margins.
    (1) Hard positive samples. Misclassified samples identified during training are labeled as hard positive samples. 
    (2) Class-wise margins. Larger margins are assigned to minority classes to ensure sufficient separation from majority classes.
    (3) Instance-wise Margins. We propose to apply adaptive margins to hard positive samples, considering both class frequency and sample difficulty.}
    \label{fig:main}
\end{figure}

Addressing class imbalance has been a focal point in long-tailed recognition (LTR) research. 
Existing methods have employed various strategies to rebalance the influence of different classes.
Re-sampling techniques, such as oversampling \cite{oversampling_byrd} and undersampling \cite{ undersampling_drummond}, adjust the occurrence of class samples to create a more balanced training set.
Re-weighting approaches \cite{cb_cui, logitadjust_menon, bs_ren} modify class weights or logit values to emphasize learning from difficult minority classes.
For instance, the label-distribution-aware margin (LDAM) loss \cite{ldam_cao} introduces larger margins for minority classes to counteract the bias towards majority classes.
Despite these advances, many methods focus primarily on class-level imbalance and often overlook variations in difficulty among individual samples within each class.
This oversight can lead to suboptimal performance on challenging instances, even within well-represented classes.

To address this gap, we propose a novel Difficulty-aware Balancing Margin (DBM) loss that considers both instance-level difficulty and class imbalance.
Unlike previous methods that primarily address class-level bias, DBM loss incorporates two key components: a class-wise margin to mitigate imbalance in class frequencies and an instance-wise margin that adapts to the difficulty of individual samples. 
By assigning additional margins to hard positive samples, our approach enhances class discriminability even more.

\Cref{fig:main} illustrates the overview of our method.
Here, we consider a binary classification problem where class A has more samples than class B.
The decision boundary determined by the classifier is denoted by the black line in \cref{fig:main}(1).
The misclassified samples, indicated by a two-line border, are identified as hard positive samples.
Our margin loss assigns a tighter decision boundary to each sample, aiming to bring sample features closer to their class centers.
First, a class-wise margin of varying sizes is applied to each sample based on its class frequency.
As a result, different decision boundaries, $b_A$ and $b_B$, are defined as shown in \cref{fig:main}(2), with the minority class experiencing a larger displacement in its decision boundary compared to the majority class.
For hard positive samples, an additional instance-wise margin is applied.
The final decision boundaries for the hard positive samples $A1$, $A2$, and $B1$ in \cref{fig:main}(3) are denoted as $b_{A1}$, $b_{A2}$ and $b_{B1}$, respectively.
Given that $A1$ exhibits a greater angular distance from the class center compared to $A2$, $A1$ is assigned a larger instance-wise margin, leading to a more shift in $b_{A1}$ relative to $b_{A2}$.
This leads to a higher loss value for difficult samples, encouraging a denser feature distribution within each class.

Our method integrates seamlessly with existing LTR techniques with negligible computational impact and demonstrates consistent performance improvements across multiple benchmarks, including CIFAR-10-LT, CIFAR-100-LT \cite{ldam_cao, decoupling_kang}, ImageNet-LT \cite{imagenetLT_liu}, and iNaturalist2018 \cite{inaturalist_van}. 
Extensive experiments validate our design choices and showcase the effectiveness and robustness of our method.

The main contributions of this paper are summarized as follows:
\begin{itemize}
    \item We propose the difficulty-aware balancing margin (DBM) loss, which effectively balances learning bias due to class imbalance and sample-level difficulty variation within a class.
    \item Our DBM loss is compatible with various existing long-tailed recognition techniques, and incurs no significant additional computational overhead.
    \item When combined with state-of-the-art methods, our approach demonstrates competitive performance on major long-tailed recognition benchmarks.
\end{itemize}

\section{Related Work}
\label{sec:relwork}

\subsection{Long-tailed Recognition}
Long-tailed recognition (LTR) has been extensively explored through multiple perspectives.
Conventional approaches focus on rebalancing the bias introduced by imbalanced class influence during training, aiming to mitigate performance degradation for minority classes.
Re-sampling methods \cite{rebalancing_buda, rebalancing_he} address the class imbalance by either undersampling majority classes \cite{undersampling_drummond, undersampling_tahir} or oversampling minority classes \cite{oversampling_byrd, dos_ando}.
Re-weighting methods \cite{cb_cui, ldam_cao, bs_ren} propose class-discriminative losses to emphasize the relative contribution of minority classes.
Logit compensation methods \cite{logitadjust_menon, gcla_li, bs_ren, marc_wang, adrw_wang} adaptively adjust logit values based on prior knowledge of the sample distribution for balancing.

Another line of LTR research focuses on enhancing the robustness of representation learning to reduce model bias.
\citeauthor{ldam_cao} \shortcite{ldam_cao} demonstrated that applying class rebalancing methods in the later stages of training can be more effective than conventional one-stage methods.
\citeauthor{decoupling_kang} \shortcite{decoupling_kang} proposed decoupling the training of the feature extractor from the classifier, which inspired later two-stage approaches \cite{bbn_zhou, mislas_zhong}.
Augmentation-based methods \cite{metasaug_li, cmo_park, cuda_ahn} aim to improve the sample diversity for tail classes.
Inspired by the robust feature representation learned through self-supervision \cite{moco_he, simclr_chen}, variants of supervised contrastive learning \cite{scl_khosla} methods have been introduced to LTR \cite{hybridscl_wang, kcl_kang, tsc_li, paco_cui, bcl_zhu}.
\citeauthor{gml_suh} \shortcite{gml_suh} integrated contrastive learning with logit compensation by introducing a Gaussian mixture likelihood loss, aiming to maximize mutual information between latent features and the ground truth labels.
They employed a teacher-student strategy to generate contrast samples using a pre-trained teacher encoder.
Ensemble-based methods \cite{ride_wang, ace_cai, ncl_li, lgla_tao} exploit the complementary knowledge from multiple experts through various incorporation methods, such as routing \cite{ride_wang} and distillation \cite{ncl_li}.

Most LTR studies assume that the tail classes are inherently more difficult to learn and therefore assign more weights to less frequent classes.
However, some recent works \cite{ala_zhao, diffnet_sinha} observed that actual class-specific performance does not always correlate with class frequency.
In response, they tried to consider classification difficulty in addition to sample distribution for re-weighting.
We share a similar motivation and introduce an adaptive margin loss that makes instance-level adjustments based on the angular distance between the positive class center and the sample feature.

\subsection{Margin Loss}
Large-margin softmax loss (L-Softmax) \cite{lsoftmax_liu} was introduced to enhance feature discrimination by encouraging intra-class compactness and inter-class separability in the embedding space.
In the domain of facial recognition, margin losses have been further explored in angular space, utilizing a cosine classifier \cite{sphereface_liu, cosface_wang, arcface_deng}.
These approaches aim to improve discriminativity by optimizing the angular separation between class centers.

Challenges arising from class imbalance have also been addressed within margin-based frameworks.
For example, face recognition methods such as fair loss \cite{fairloss_liu} and AdaptiveFace \cite{adaptiveface_liu}, and label-distribution-aware margin (LDAM) loss \cite{ldam_cao} for LTR adaptively adjust class-wise margin values or sampling frequencies to mitigate bias. 
LDAM loss assigns larger margins to minority classes by explicitly incorporating class distribution priors, which helps counteract the imbalance. 
However, LDAM loss applies a uniform margin to all samples within a class, without accounting for variations in sample difficulty.
In contrast, we propose a difficulty-aware balancing margin (DBM) loss, which introduces the consideration of instance difficulty to assign even larger margins to challenging samples.
By adapting the margin based on the angular distance between the positive class center and the sample feature, DBM loss provides a more refined approach to margin adjustment, effectively addressing both class imbalance and individual sample difficulty.

\section{Proposed Method}
\label{sec:methods}

\subsection{Preliminaries}
\subsubsection{Loss functions for Long-tailed Recognition.}
The cross-entropy loss with softmax function is defined as:
\begin{equation}
    L_{\text{CE}} = -\log{{e^{\psi_{y}(x)}} \over \sum_{i}e^{\psi_{i}(x)}}
    \text{.}
    \label{eq:ce_loss}
\end{equation}
Here, $\psi_i(x)$ represents the logit function of the $i$-th class for sample $x$, which belongs to the class of index $y$.
For models that utilize a linear classifier, the logit function is given by $\psi_{i}(x) = W^\top_{i} f(x)+ b_i$, where $f(x)$ denotes the feature representation of sample $x$, and
$W_i$ and $b_i$ represent the weight and bias of the linear classifier for the $i$-th class, respectively.
Alternatively, a cosine classifier embeds features and class centers in an L2-normalized space, with logits determined by the angular distance between sample features and class centers. 
Specifically, 
\begin{equation}
\psi_{i}(x) = s {W^\top_{i} f(x) \over \|W_{i}\| \|f(x)\|} = s \cos{\theta_i}, 
\label{eq:logit_function}
\end{equation}
where $s$ is the scaling factor and $\theta_i$ denotes the angular distance between $W_i$ and $f(x)$.

In long-tailed recognition (LTR), re-weighting methods address class imbalance by incorporating class frequency $n_i$ into the loss functions.
Variants of cross-entropy loss include the class-balanced (CB) loss \cite{cb_cui} and balanced softmax (BS) \cite{bs_ren}.
The class balanced loss $L_{\text{CB}}$ is formulated as: 
\begin{equation}
    L_{\text{CB}} = -{1-\beta \over 1-\beta^{n_y}}\log{{e^{\psi_{y}(x)}} \over \sum_{i}e^{\psi_{i}(x)}}
    \text{,}
\label{eq:cb_loss}
\end{equation}
introducing a class-wise weight determined by the effective number of samples given a hyperparameter $\beta$.
The balanced softmax loss $L_{\text{BS}}$ is: 
\begin{equation}
    L_{\text{BS}} = -\log{{e^{\psi_{y}(x)+\log{p_y}}} \over \sum_{i}e^{\psi_{i}(x)+\log{p_i}}}
    \text{,}
\label{eq:bs_loss}
\end{equation}
where $p_i$ represents the sample proportion of the $i$-th class over all classes, i.e., $ p_i = {n_i / \sum_{j}n_j}$.
The balanced softmax loss is widely adopted in later LTR studies, such as balanced contrastive learning (BCL) \cite{bcl_zhu} and nested collaborative learning (NCL) \cite{ncl_li}.

\subsubsection{Margin-based Variants of Cross-entropy Loss.}
Margin losses introduce a specialized logit function associated with a margin for the positive class.
A margin-based cross-entropy loss $L_{m}$ can be generally formulated as:
\begin{equation}
    L_{m} = -\log{e^{s \psi^m_y(\theta_{y})} \over e^{s \psi^m_y(\theta_{y})} + \sum_{i \neq y}e^{s\cos\theta_{i}}}
    \text{,}
    \label{eq:margin_loss}
\end{equation}
where $s\psi^m_y(\theta_{y})$ denotes the logit function for the positive class incorporating the margin.
If $\psi^m_y(\theta_y)$ adopts no margin, i.e., $\psi^m_y(\theta_y) = \cos \theta_y$, $L_m$ is equivalent to $L_{\text{CE}}$.

\begin{table}[h]
    \centering
    \small
    \begin{tabularx}{0.4\textwidth}{p{0.22\textwidth}|Y}
    \toprule
    Methods & $\psi^m_y(\theta_{y})$\\
    \midrule
    {\small SphereFace \cite{sphereface_liu}}    & $\cos{(m\theta_{y})}$ \\
    {\small CosFace \cite{cosface_wang}}         & $\cos{\theta_{y}}-m$ \\
    {\small ArcFace \cite{arcface_deng}}         & $\cos{(\theta_{y}+m)}$ \\
    {\small LDAM \cite{ldam_cao}}                & $\cos{\theta_{y}}-{m n^{-{1/4}}_{y}}$ \\
    
    \bottomrule
    \end{tabularx}
    \caption{$\psi^m_y(\theta_{y})$ used in different margin losses.}
    \label{table:margin-loss}
    \setlength{\tabcolsep}{1mm}
\end{table}

\Cref{table:margin-loss} provides a summary of various margin-based loss functions and their respective logit formulations.
Traditional margin losses \cite{sphereface_liu, cosface_wang, arcface_deng} apply a constant margin for all classes.
CosFace \cite{cosface_wang} applies a margin to the measured cosine similarity, while ArcFace \cite{arcface_deng} directly adjusts the angular distance.
LDAM loss \cite{ldam_cao} follows a similar formulation to CosFace, subtracting a margin that varies with class frequency from the cosine similarity to address the  class imbalance problem.

\subsection{Difficulty-aware Balancing Margin Loss}
\label{subsec:DBM}

Our difficulty-aware balancing margin (DBM) loss comprises two components: a class-wise margin and an instance-wise margin.
By integrating these two elements, we address both the  bias from class imbalance and the variation in instance difficulty within a class.
Following prior works \cite{ihem_Xiao, hamface_Li}, we apply the instance-wise margin specifically to hard positive samples.
\Cref{fig:margin} illustrates the margins determined by class frequency and angular distance.
Detailed mathematical descriptions of each component are provided below.

\subsubsection{Class-wise Margin.}
The class-wise margin $m_{C}$ is defined as:
\begin{equation}
    {m}_{C} = K \rho_y ^{-\tau}
    \text{.}
\end{equation}
Here, $\rho_y = n_y / n_{\min}$ represents the ratio of the number of samples in class $y$ to the number in the least frequent class. 
The parameter $\tau$ controls the extent of the margin difference across classes, while $K$ scales the margin.
As illustrated in \cref{fig:margin_cls}, $m_C$ is solely based on the class frequency ratio $\rho_y$.
By scaling inversely with $\rho_y$, minority classes receive a larger margin compared to majority classes, ensuring the least frequent class obtains the maximum margin of $K$.
This helps mitigate performance degradation for minority classes.
We have found that setting $\tau=1$ is effective for our approach.

\subsubsection{Instance-wise Margin.}
The instance-wise margin addresses varying sample-level difficulties.
Samples with lower positive logit values are more prone to misclassification.
For our cosine classifier, difficult samples are those whose feature representations are farther from the positive class center in the hypersphere.
We quantify the instance difficulty $d_I$ via following equation:
\begin{equation}
    d_{I} = {1 - \cos\theta_{y} \over 2}
    \text{.}
\end{equation}
Here, $d_I$ is determined by the angular distance between the feature representation of the sample and the positive class center $\theta_y$.
A sample with its feature representation exactly at the class center has $d_I = 0$, while a sample with the feature representation at the maximum distance ($\theta_y = \pi$) has $d_I = 1$.

\begin{figure}[t!]
    \centering   
    \begin{subfigure}[b]{0.49\linewidth}
        \centering
        \includegraphics[width=\textwidth]{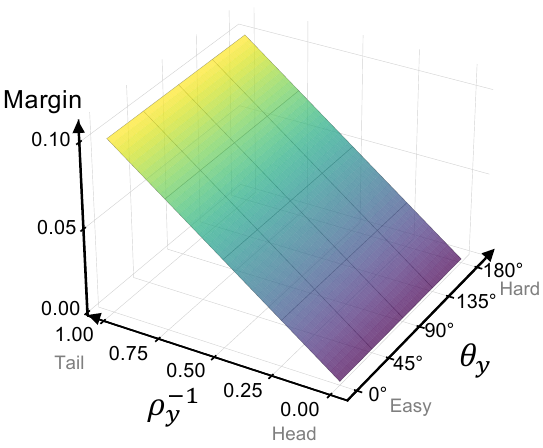}
        \caption{Class-wise margin.}
        \label{fig:margin_cls}
    \end{subfigure}
    \hfill
    \begin{subfigure}[b]{0.49\linewidth}
        \centering
        \includegraphics[width=\textwidth]{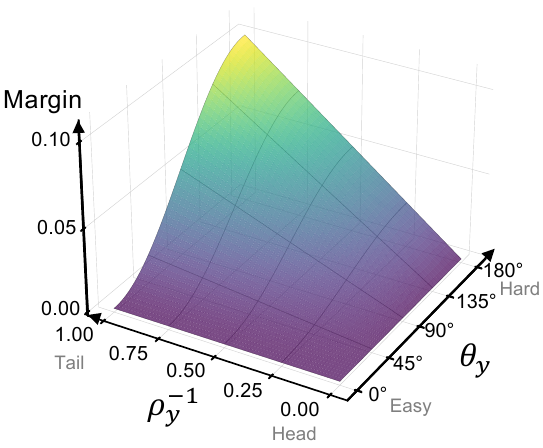}
        \caption{Instance-wise margin.}
        \label{fig:margin_inst}
    \end{subfigure}
    \caption{Margins for $K=0.1$ and $\tau=1$. Less frequent classes have larger class-wise margin, and more difficult samples have larger instance-wise margin.}
    \label{fig:margin} 
\end{figure}

The instance-wise margin $m_{I}$ is given by:
\begin{equation}
    \label{eq:inst_margin}
    {m}_{I} = m_{C} \cdot d_{I}.
\end{equation}
As illustrated in \cref{fig:margin_inst}, this margin is determined by both $\rho_y$ and $\theta_y$, encouraging difficult and less-frequent samples to move more aggressively towards the positive class center.

\subsubsection{Loss formulation.}
Our DBM loss modifies the angular distance by incorporating both margins, similar to the ArcFace approach.
Specifically, our logit function for the positive class is formulated as:
\begin{equation}
\small
    s \psi^{dbm}_y(\theta_y) = s\cos(\theta_y + m_C + \mathbb{1}[\underset{i}{\mathrm{argmin}}(\{\theta_{i}\}_{i=1}^N) \neq y] m_I)
    \text{,}
    \label{eq:dbm_logit_function}
\end{equation}
where $\mathbb{1}[\cdot]$ is an indicator function for applying the instance-wise margin only to hard positive samples.
By substituting this logit function into \cref{eq:margin_loss}, we derive the difficulty-aware balancing margin cross-entropy (DBM-CE) loss.

The DBM loss can be easily integrated with various existing LTR methods. 
For example, it can be combined with the class-balanced loss introduced in \cref{eq:cb_loss} as follows:
\begin{equation}
\small
\label{eq:DBM_CB}
    L_{\text{DBM-CB}} = -{1-\beta \over 1-\beta^{n_y}}\log{{e^{s \psi^{dbm}_y(\theta_y)}} \over e^{s \psi^{dbm}_y(\theta_y)}+\sum_{i \neq y}e^{s\cos\theta_{i}}}
    \text{.}
\end{equation}
Similarly, DBM-BS can be derived as:
\begin{equation}
\small
\label{eq:DBM_BS}
    L_{\text{DBM-BS}} = -\log{e^{s\psi^{dbm}_y(\theta_y) + \log p_y}
    \over e^{s \psi^{dbm}_y(\theta_y) + \log p_y}+\sum_{i \neq y}e^{s\cos\theta_{i}+ \log p_i}}
    \text{,}
\end{equation}
reformulating the original balanced softmax loss described in \cref{eq:bs_loss}.
Note that our method requires adjusting the classifier from a linear to a cosine classifier.

Moreover, Our method is highly versatile and can be incorporated with a range of other LTR techniques.
We demonstrate this versatility with various configurations of our method, including DBM-DRW, DBM-BCL, DBM-GML, and DBM-NCL.
DRW, or deferred re-weighting \cite{ldam_cao}, integrates class-balanced loss into the training process at a later stage, allowing DBM-DRW to be implemented by applying $L_{\text{DBM-CE}}$ and $L_{\text{DBM-CB}}$ sequentially according to the scheduling policy.
Similarly, methods like BCL \cite{bcl_zhu}, GML \cite{gml_suh} and NCL \cite{ncl_li}, which originally use balanced softmax loss, can incorporate our approach by substituting the classification loss with $L_{\text{DBM-BS}}$.

The integration of DBM loss into existing models does not incur significant additional computational complexity. 
The class-wise margin $m_C$ is determined in advance based on the known sample distribution, ensuring that this computation does not affect the training time. 
The instance-wise margin $m_I$ is computed during the logit calculation, leveraging the angular distance $\theta_y$ that is already part of the model's forward pass. 
This design ensures that DBM can be incorporated into existing frameworks without introducing substantial overhead.

\section{Experiments}
\label{sec:exp}
\subsection{Datasets}
To evaluate the performance of our proposed method, we conducted experiments on four benchmark long-tailed datasets.
The imbalance factor $\rho$ of each dataset is defined as the ratio of training instances between the largest and smallest classes, i.e., $\rho = n_{\max}/n_{min}$, following previous works \cite{ldam_cao,decoupling_kang}.

\sisetup{detect-weight,mode=text}
\newcommand{\B}{\fontseries{b}\selectfont}

\renewrobustcmd{\boldmath}{}

\begin{table*}[!t]
\centering
\begin{tabularx}{0.9\textwidth}{l|YYY|YYY|YYY}
\toprule
\multirow{3}{*}{Method} & \multicolumn{3}{c|}{CIFAR-10-LT} & \multicolumn{6}{c}{CIFAR-100-LT}  \\
\cmidrule{2-4} \cmidrule{5-10}
&\multicolumn{3}{c|}{Imb. Factor} &\multicolumn{3}{c|}{Imb. Factor}  & \multicolumn{3}{c}{Statistics (IF 100)} \\
& 100 & 50 & 10 & 100 & 50 & 10 & Many & Med. & Few \\
\midrule
CE & 78.48 & 82.73 & 89.91 & 44.60 & 48.75 & 61.98 & 73.03 & 45.37 & 10.53 \\
LDAM \cite{ldam_cao} & 79.92 & 83.84  & 90.54 & 45.25 & 50.16 & 62.86 & \B{75.31} & 44.00 & 11.63 \\
\rowcolor{LightCyan}
DBM-CE & \B{80.84} & \B{84.12} & \B{90.95} & \B{46.53} & \B{51.13} & \B{63.18} & 73.89 & \B{46.17} & \B{15.03} \\
\midrule
CE-DRW \cite{ldam_cao} & 82.24 & 85.05 & 90.94 & 48.28 & 53.89 & 64.25 & 65.89 & 50.74 & 24.87\\
LDAM-DRW \cite{ldam_cao} & 82.60 & 85.36 & 91.22 & 48.99 & 54.27 & 64.58 & \B{66.09} & 50.83 & 26.90 \\
\rowcolor{LightCyan}
DBM-DRW & \B{82.82} & \B{85.83} & \B{91.55} & \B{49.41} & \B{54.69} & \B{64.75} & 63.23 & \B{52.66} & \B{29.50} \\
\midrule
BS \cite{bs_ren} & 83.57 & 86.45 & 91.26 & 
49.35 & 54.79 & 63.93 & 65.77 & 50.14 & 29.27 \\
\rowcolor{LightCyan}
DBM-BS & \B{84.60} & \B{87.06} & \B{91.42} & \B{51.30} & \B{55.84} & \B{65.22} & \B{67.29} & \B{50.80} & \B{33.23}\\
\midrule
BCL \cite{bcl_zhu} & 82.95 & 86.76 & 91.57& 50.23 & 55.35 & 64.98 & 67.14 & 51.31 & 29.23\\
\rowcolor{LightCyan}
DBM-BCL & \B{84.60} & \B{87.16} & \B{91.69} &  \B{51.66} & \B{55.98} & \B{65.25} & \B{67.91} & \B{51.91} & \B{32.40} \\
\midrule
GML \cite{gml_suh} & 85.19 & 88.07 & 92.11 & 53.12 & 58.17 & 66.93 & 71.60 & 54.57 & 28.20 \\ 
\rowcolor{LightCyan}
DBM-GML & \B{85.30} & \B{88.35} & \B{92.59} & \B{53.70} & \B{58.41} & \B{67.15} & \B{72.34} & \B{54.89} & \B{30.57} \\
\midrule
NCL \cite{ncl_li} & 87.37 & 89.89 & 93.15 & 56.68 & 61.65 & 69.46 & \B{73.94} & 56.97 & 36.20 \\
\rowcolor{LightCyan}
DBM-NCL & \B{87.53} & \B{89.90} & \B{93.19} & \B{57.48} & \B{62.01} & \B{69.75} & 71.49 & \B{59.06} & \B{39.30} \\
\bottomrule
\end{tabularx}
\caption{Top-1 accuracy (\%) of ResNet-32 on CIFAR-10-LT and CIFAR-100-LT with the imbalance factor (IF) of 100, 50, and 10.}
\label{table:cifar100}
\setlength{\tabcolsep}{1mm}
\end{table*}

\subsubsection{Long-tailed CIFAR-10 and CIFAR-100.}
We sampled long-tailed CIFAR datasets from the original CIFAR-10 and CIFAR-100 \cite{cifar_krizhevsky} datasets with imbalance factors of 10, 50, and 100 using an exponential down-sampling profile outlined in \cite{ldam_cao, cb_cui}.
Evaluations were performed on the original balanced test sets.

\subsubsection{ImageNet-LT.} 
ImageNet-LT \cite{imagenetLT_liu} is a long-tailed version of ImageNet-1K \cite{imagenet_deng}, sampled from a Pareto distribution with $\alpha = 6$.
It comprises 1,000 categories and 115.8K training images, with an imbalanced factor of $\rho=1280/5$.

\subsubsection{iNaturalist2018.} 
The iNaturalist2018 dataset \cite{inaturalist_van} is a large-scale real-world dataset that features a highly long-tailed distribution with an imbalance factor of $\rho = 1000/2$.
It includes approximately 437K training images and 24.4K validation images gathered from 8,142 fine-grained species classes in the wild.

\subsection{Implementation Details}
For the CIFAR-10-LT and CIFAR-100-LT datasets, we integrated our method with several existing approaches including:

\begin{itemize}
    \item[(1)] vanilla cross-entropy (CE)
    \item[(2)] CE-DRW \cite{ldam_cao}, a two-stage training method applying CB loss \cite{cb_cui}.
    \item[(3)] BS \cite{bs_ren}, a re-weighting method.
    \item[(4)] BCL \cite{bcl_zhu}, a supervised contrastive learning-based method. 
    \item[(5)] GML \cite{gml_suh}, a mutual information maximization method.
    \item[(6)] NCL \cite{ncl_li}, an ensemble-based method.
\end{itemize}
We ensured a fair comparison by evaluating our models under identical experimental conditions.
All models utilized ResNet-32 \cite{resnet_he} as the backbone network, while ResNet56 was employed as the teacher network for GML.
The SGD optimizer with a momentum of 0.9 and weight decay of $2\times10^{-4}$ was employed, along with a learning rate warm-up for the first five epochs and a cosine annealing scheduler for gradual decay.
Data augmentation strategies included Cutout \cite{cutout_devries} and AutoAugment \cite{autoaug_cubuk}. 
For BCL, we used an initial learning rate of 0.15 and a batch size of 256.
For all other methods, we used an initial learning rate of 0.1 and a batch size of 64.
Training was conducted for 200 epochs for most methods, except for NCL, which was trained for 400 epochs.
In the case of DRW, class-balanced loss is introduced after 160 epochs.
We used a scaling factor $s=32$ for all our experiments, and tuned the hyperparameter for margin scaling $K$ within the range 0.1 to 0.3, adjusting it based on datasets and baselines.

For larger datasets, our method was integrated into BCL and GML.
NCL was excluded from this comparison due to its extensive training requirements of 400 epochs.
For ImageNet-LT, we utilized ResNet-50 and ResNeXt-50 \cite{resnext_xie} as backbones and trained them for 90 epochs.
For iNaturalist2018, we employed ResNet-50 and trained for 100 epochs.
In both benchmarks, we set the scaling factor $s$ to 30 and the margin scaling hyperparameter $K$ to 0.1.
Further details are in the supplementary materials.

\subsection{Experimental Results}
\subsubsection{Long-tailed CIFAR.}
\Cref{table:cifar100} presents the experimental results for CIFAR-10-LT and CIFAR-100-LT.
For CIFAR-100-LT with an imbalance factor of 100, we report the accuracy across three groups of classes: `Many ($>$ 100 shots),' `Medium (20$\sim$100 shots),' and `Few ($<$ 20 shots).'
To ensure fairness, we have reproduced the performance of each previous method and provided these results in the corresponding cells.
Methods incorporating DBM loss are highlighted in cyan.

The results demonstrate that DBM consistently provides a significant performance improvement over baseline methods. 
When applied to CE and CE-DRW, our method achieves superior enhancement compared to LDAM and LDAM-DRW, which solely introduces a class-wise margin.
Notably, DBM-BS surpasses BCL, indicating a substantial performance boost without the additional complexity introduced by BCL's contrastive learning branch.
Although some algorithms show a slight decrease in accuracy for the `Many' group compared to the baseline, our method achieves a notable increase in accuracy for the `Medium' and `Few' groups, demonstrating its effectiveness in mitigating performance bias.

\begin{table}[t]
\centering
\setlength{\tabcolsep}{1mm}
\begin{tabularx}{0.47\textwidth}{l|YY}
\toprule
Method & R50 & RX50 \\
\midrule
{\small CE$^\dagger$} & 41.6 & 44.4 \\
{\small $\tau$-norm \cite{decoupling_kang}} & 46.7 & 49.4 \\
{\small cRT \cite{decoupling_kang}} & 47.3 & 49.6 \\
{\small LWS \cite{decoupling_kang}} & 47.7 & 49.9 \\
{\small LDAM-DRW$^\ddagger$ \cite{ldam_cao}} & 49.8 & $-$ \\
{\small CE-DRW$^\ddagger$ \cite{ldam_cao}} & 50.1 & $-$ \\
{\small BS$^\ddagger$ \cite{bs_ren}} & 50.9 & $-$ \\
{\small ALA Loss \cite{ala_zhao}} & 52.4 & 53.3 \\
{\small DisAlign \cite{disalign_zhang}} & 52.9 & 53.4 \\
{\small Difficulty-Net \cite{diffnet_sinha}} & 54.0 & $-$ \\
{\small RIDE (3 experts) \cite{ride_wang}} & 54.9 & 56.4 \\
{\small BCL \cite{bcl_zhu}} & 56.0 & 56.7 \\
{\small GML \cite{gml_suh}} & $-$ & 58.3 \\
\midrule
{\small DBM-BCL} & 56.3 & 57.4 \\
{\small DBM-GML} & \B{57.4} & \B{58.6} \\
\bottomrule
\end{tabularx}
\caption{Top-1 accuracy (\%) of ResNet-50 and ResNeXt-50 on ImageNet-LT. $^\dagger$ and $^\ddagger$ denotes results borrowed from \citeauthor{decoupling_kang} (\citeyear{decoupling_kang}) and \citeauthor{cmo_park} (\citeyear{cmo_park}), respectively.}
\label{table:imgnet}\setlength{\tabcolsep}{1mm}
\end{table}

\begin{table}[t]
\centering
\begin{tabularx}{0.47\textwidth}{p{0.23\textwidth}|YYY|Y}
\toprule
Methods & \small{Many} & \small{Med.} & \small{Few} & All\\
\midrule
{\small CE$^\dagger$} & \B{73.9} & 63.5 & 55.5 & 61.0 \\
{\small $\tau$-norm \cite{decoupling_kang}}  & 65.6 & 65.3 & 65.9 & 65.6 \\
{\small cRT \cite{decoupling_kang}} & 69.0 & 66.0 & 63.2 & 65.2 \\
{\small LWS \cite{decoupling_kang}} & 65.0 & 66.3 & 65.5 & 65.9 \\
{\small LDAM-DRW$^\dagger$ \cite{ldam_cao}} & $-$ & $-$ & $-$ & 66.1 \\
{\small CE-DRW$^\ddagger$ \cite{ldam_cao}} & 68.2 & 67.3 & 66.4 & 67.0 \\
{\small BS$^\ddagger$ \cite{bs_ren}} & 65.5 & 67.5 & 67.5 & 67.2 \\
{\small DisAlign \cite{disalign_zhang}} & 61.6 & 70.8 & 69.9 & 69.5 \\
{\small RIDE \cite{ride_wang}} & 70.2 & 71.3 & 71.7 & 71.4 \\
{\small BCL$^\star$ \cite{bcl_zhu}} & 68.2 & 71.3 & 71.3 & 71.0 \\
{\small GML$^\star$ \cite{gml_suh}} & 70.7 & \B{72.3} & 71.5 & 71.2 \\
\midrule
{\small DBM-BCL}  & 65.6 & 71.8 & 73.8 & 71.9 \\
{\small DBM-GML}  & 66.9 & 71.9 & \B{73.6} & \B{72.0} \\
\bottomrule
\end{tabularx}
\caption{Top-1 accuracy (\%) of ResNet-50 on iNaturalist2018. $^\dagger$ and $^\ddagger$ denotes results borrowed from \citeauthor{bbn_zhou} (\citeyear{bbn_zhou}) and \citeauthor{cuda_ahn} (\citeyear{cuda_ahn}), respectively. $^\star$ denotes reproduced results with the official code. RIDE (2 experts) \cite{ride_wang} was trained for 100 epochs.}
\label{table:iNat}\setlength{\tabcolsep}{1mm}
\end{table}

\subsubsection{ImageNet-LT and iNaturalist2018.}
\Cref{table:imgnet} shows the performance of DBM-BCL and DBM-GML compared to the existing methods on the ImageNet-LT dataset.
We report overall accuracy using ResNet-50 and ResNeXt-50 backbones.
For a fair comparison, we evaluated our method against existing works that reported the performance after 90 epochs of training.
DBM-BCL outperforms the baseline BCL, with an overall accuracy improvements of 0.3\%p and 0.7\%p for the ResNet-50 and ResNeXt-50 backbones, respectively.
Although GML did not report results for the ResNet-50 model, DBM-GML demonstrates an improved performance of 0.3\%p for the ResNeXt-50 backbone.

\Cref{table:iNat} displays the performance comparisons on the iNaturalist2018 dataset.
We report overall accuracy and the accuracy of ‘Many,’ ‘Medium,’ and ‘Few’ groups in our experiment.
To ensure a fair comparison, we excluded methods that involve extensive additional training \cite{paco_cui, ncl_li}.
Since BCL and GML did not report accuracy for each group, we re-implemented their results using their official code.
DBM-BCL and DBM-GML achieve improvements of 0.9\%p and 0.8\%p in overall accuracy, respectively, surpassing the performances of existing methods.

\subsection{Analysis}
\label{sec:Analysis}

In this section, we analyze the components of the DBM loss to evaluate their contributions to performance improvement. 
We also investigate the impact of different hyperparameters on the method's effectiveness.
Additionally, we illustrate how the introduced margin enhances intra-class compactness and inter-class separability, thus improving classification performance.
All experiments for analysis were conducted on CIFAR-100-LT with an imbalance factor of 100.

\begin{table}[t]
\centering
\begin{tabularx}{0.47\textwidth}{YYY|YY}
\toprule
Cosine & $m_C$ & $m_I$ & CE & BS \\
\midrule
& & & 44.60 & 49.35\\
\checkmark & & & 44.29 & 49.84\\
\checkmark & \checkmark & & 45.85 & 50.61 \\
\checkmark & \checkmark & P & 46.38 & 50.93 \\
\checkmark & \checkmark & HP(ours) & \B{46.53} & \B{51.30} \\

\bottomrule
\end{tabularx}
\caption{Ablation study for the components of DBM loss. `Cosine' denotes replacing the linear classifier with cosine classifier. $m_C$ and $m_I$ denote class-wise and instance-wise margin, respectively. P and HP represent the cases where instance-wise margin is applied to all positive samples and hard positive samples, respectively.}
\label{table:components}\setlength{\tabcolsep}{1mm}
\end{table}

\subsubsection{Component Analysis.}
\Cref{table:components} presents the results of our ablation study, which examines the impact of class-wise and instance-wise margins.
We integrated these components into two baseline methods: CE and BS \cite{bs_ren}.
Our findings reveal that using a cosine classifier alone does not significantly improve performance.
However, incorporating a class-wise margin leads to notable gains.
Adding the instance-wise margin results in an additional performance increase of approximately 0.7\%p for both loss functions.

We also observed differences in performance based on the way the instance-wise margin is applied.
Specifically, the `hard positive (HP)' strategy, where the margin is applied only to misclassified positive samples, yields better results compared to the `positive (P)' strategy, which applies the margin to all positive samples.
This indicates that focusing on the difficulty of hard positive samples only rather than all positive samples improves performance more effectively.

\subsubsection{Impacts of Hyperparameters.}
\begin{figure}[!t]
    \centering
    \includegraphics[width=\columnwidth]{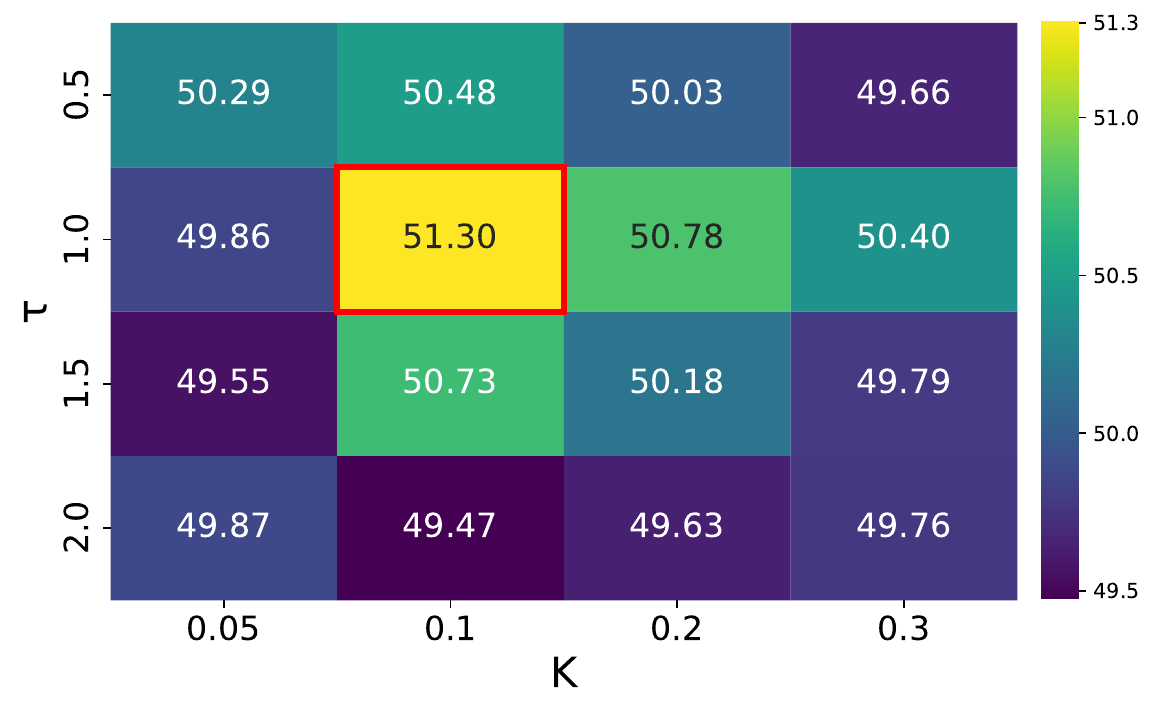}
    \caption{Analysis for effects of hyperparameters $\tau$ and $K$. For all cases, DBM-BS outperforms the baseline BS \cite{bs_ren} performance of 49.35\%.}
    \label{fig:hp_tuning}
\end{figure}

\Cref{fig:hp_tuning} illustrates the effects of various hyperparameters on the performance of DBM-BS.
We analyze the impact of $\tau$ and $K$, which are critical parameters in our method. 
The results show that while variations in these hyperparameters cause slight performance differences, DBM consistently outperforms the baseline across all settings.

Based on our observation, we fixed $\tau$ at 1.0 throughout all experiments on the long-tailed benchmarks.
The optimal value for $K$ varies depending on the method, but setting $K=0.1$ generally yields satisfactory results.

\subsubsection{Intra-class Compactness and Inter-class Separability.}
\begin{figure}[!t]
    \centering
    \includegraphics[width=\linewidth]{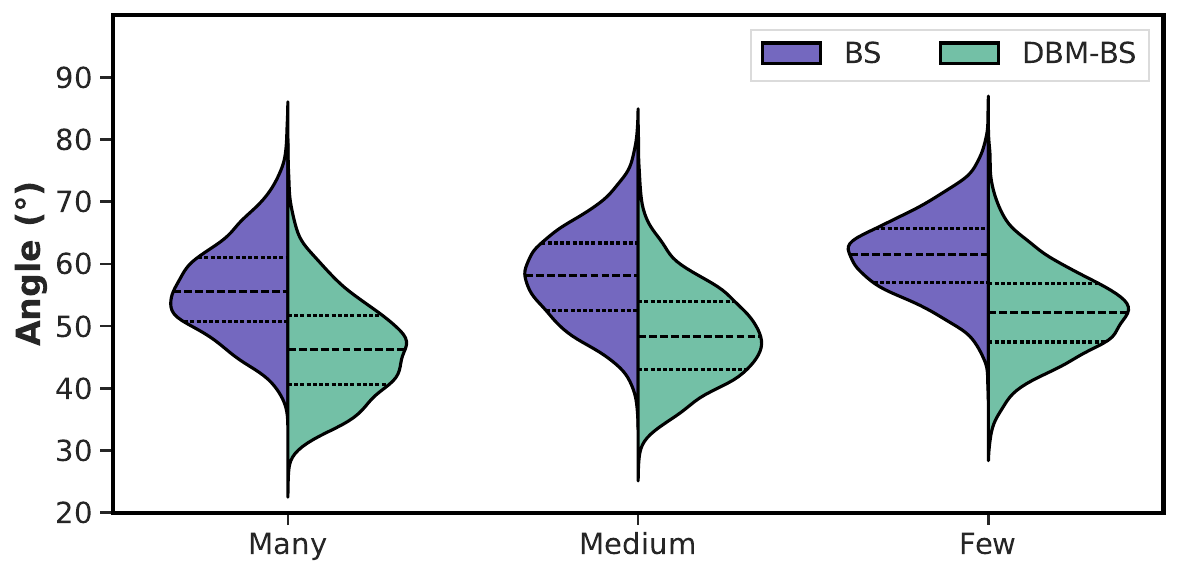}
    \caption{Comparison of BS \cite{bs_ren} and DBM-BS of the distribution of angular distance between sample features and their positive class centers for `Many', `Medium', and `Few' groups. Dashed horizontal lines denote the quartiles.}
    \label{fig:intra-comp}
\end{figure}

We apply instance-wise margins to bring hard positive samples closer to their respective class centers, aiming to enhance intra-class compactness. 
\Cref{fig:intra-comp} compares the distribution of angular distances between sample features and their positive class centers for the `Many', `Medium', and `Few' groups in BS and DBM-BS.
DBM-BS shows a reduction in the mean angular distance of approximately $10^{\circ}$ across all groups, indicating enhanced intra-class compactness. 
This suggests that DBM improves the alignment of sample features with their respective class centers, which may contribute to better performance in classification tasks.

To evaluate inter-class separability, we use the Fisher criterion from Fisher's linear discriminant analysis (LDA) \cite{lda_fisher} as a metric to measure the distance between feature distributions of different classes.
LDA aims to find a projection vector $W$ that maximizes the separation between classes by projecting the data onto a new axis where the classes are most distinguishable.
The Fisher criterion is used to determine the optimal $W$ that maximizes the ratio of the between-class variance to the within-class variance.

The Fisher criterion is defined as:
\begin{equation}
\label{eq:fisher}
    J(W_{ij}) = {({\mu_i - \mu_j)^2} \over {\sigma^{2}_{i} + \sigma^{2}_{j}}}
    \text{,}
\end{equation}
where $W$ is the projection vector, and $\mu_k$ and $\sigma_k^2$ denote the mean and variance of the projected feature distribution for the $k$-th class, respectively.
The objective is to find $W_{ij}$ such that the means of the projected classes $\mu_i$ and $\mu_j$ are as far apart as possible while the variances $\sigma^2_i$ and $\sigma^2_j$ are minimized.
A higher value of the Fisher criterion $J(W_{ij})$ indicates greater separability between the two classes.

After calculating the optimal projection vectors for all class pairs, we define the separability of a class $S_i$ as:
\begin{equation}
\label{eq:separability}
    S_i = {1 \over {C-1}}\sum^C_{j=1, j\neq i}J(W_{ij})
\end{equation} 
where $C$ is the number of classes.
\Cref{table:inter-sep} presents the separability for `Many,' `Medium,' `Few,' and `All' groups.
Our observations confirm that DBM enhances inter-class separability across all groups, leading to improved overall classification performance.

\begin{table}[t]
\centering
\begin{tabularx}{0.47\textwidth}{p{0.18\textwidth}|YYY|Y}
\toprule
Method & Many & Med. & Few & All \\
\midrule
BS \cite{bs_ren}      & 6.02 & 5.98 & 5.65 & 5.89 \\
DBM-BS  & \B{6.21} & \B{6.26} & \B{5.95} & \B{6.15} \\
\bottomrule
\end{tabularx}
\caption{Analysis for inter-class separability. A larger value indicates better separability.}
\label{table:inter-sep}\setlength{\tabcolsep}{1mm}
\end{table}

\section{Conclusion}
\label{sec:Conclusion}
In this work, we propose a difficulty-aware balancing margin (DBM) loss, a novel approach designed to address class-level imbalance and instance-level difficulty variations in long-tailed datasets. 
The DBM loss incorporates a class-wise margin to mitigate the performance degradation caused by class imbalance and a instance-wise margin to enhance class discriminability by more effectively aligning misclassified samples with their corresponding class centers.
Our method integrates effortlessly with existing long-tailed recognition techniques and consistently improves performance across benchmarks.
We comprehensively evaluated our method on the long-tailed CIFAR, ImageNet-LT, and iNaturalist2018 datasets, and demonstrated its effectiveness through extensive experiments.

\section*{Acknowledgments}
This work was supported by the National Research Foundation of Korea (NRF) grant funded by the Korean government (MSIT) (NRF-2018R1A5A7025409).

\bibliography{aaai25}

\makeatletter
\NewDocumentCommand{\MakeTitleInner}{+m}{
    \newpage%
    \null%
    \vskip 0em%
    \begin{center}%
        \let \footnote \thanks
        {\LARGE \textbf{#1} \par}
    \end{center}%
    \par
    \vskip 3.5em%
}
\NewDocumentCommand{\MakeTitle}{+m}{%
    \begingroup
        \if@twocolumn
            \ifnum \col@number=\@ne
                \MakeTitleInner{#1}
            \else
                \twocolumn[\MakeTitleInner{#1}]%
            \fi
        \else
            \newpage
            \global\@topnum\z@   
            \MakeTitleInner{#1}
        \fi
        \thispagestyle{plain}\@thanks
    \endgroup
}
\makeatother

\MakeTitle{Supplementary Materials for \\
``Difficulty-aware Balancing Margin Loss for Long-tailed Recognition''}
\setcounter{table}{0}

\section*{Implementation Details}
This section outlines the implementation details for our experiments on the ImageNet-LT \cite{imagenetLT_liu} and iNaturalist2018 \cite{inaturalist_van} datasets.
For experiments combining our DBM loss with BCL \cite{bcl_zhu} and GML \cite{gml_suh}, we applied RandAugment \cite{randaug_cubuk} augmentation strategy for the classifier training and SimAugment \cite{simclr_chen} for contrastive learning.
In the GML setup, we used ResNet-152 \cite{resnet_he} as the teacher network for the ResNet-50 student network, and ResNeXt-101 \cite{resnext_xie} as for the ResNeXt-50.
The optimizer parameters are summarized in \cref{tab:hp_detail}.

\begin{table}[h]
\centering
\setlength{\tabcolsep}{1mm}
\begin{tabularx}{\linewidth}{p{0.28\linewidth}|Y|Y|Y|Y}
\toprule
\multirow{2}{*}{Dataset} & \multirow{2}{*}{Method} & Batch & Learning & Weight \\
& & size & rate & decay \\
\midrule
\multirow{2}{*}{ImageNet-LT} & BCL & 256 
& 0.1 & 5e-4 \\ 
 &GML & 128 & 0.05 & 5e-4 \\ 
 \midrule
\multirow{2}{*}{iNaturalist2018}  & BCL & 256 & 0.2 & 1e-4 \\ 
 & GML & 128 & 0.02 & 2e-4 \\ 
 \bottomrule
\end{tabularx}
\caption{The hyperparameters for BCL and GML methods on different datasets.}
\label{tab:hp_detail}
\setlength{\tabcolsep}{1mm}
\end{table}

\section*{Additional Evaluations}

\subsection*{Long-tailed CIFAR}

In the main paper, we demonstrated the performance enhancement of our method when applied to various LTR techniques.
Here, we extend our evaluation by comparing our method with additional approaches not include in the main paper.

\paragraph*{Comparison with Difficulty-based Approach.}
Yu et al. (2022) proposed an instance-level re-sampling method based on the difficulty of each instance, determined by its learning speed. 
Since we share similar motivations for considering instance-level scaling via difficulty, we compare the performance of both techniques in this section.
Following their experiment settings, we evaluate the performance of ResNet-32 on CIFAR-10-LT and CIFAR-100-LT with imbalance factor of 100, 50, and 20. 
In these experiments, we used a different scheduler, decaying the learning rate at epoch 160 and 180 with a step size of 0.1.
Advanced augmentation strategies, such as Cutout \cite{cutout_devries} and AutoAugment \cite{autoaug_cubuk}, were not utilized in these settings.

\Cref{table:inst_difficulty} shows the comparison results. 
Despite differences in experimental setups, DBM-CE consistently outperforms CE across all imbalance factors.
Moreover, DBM-BS significantly surpasses the performance of the previous work \cite{instancedifficulty_yu}, achieving the highest performance across all imbalance factors.

\begin{table}[t]
\centering
\begin{tabularx}{0.48\textwidth}{p{0.12\textwidth}|YYY|YYY}
\toprule
\multirow{3}{*}{Method} & \multicolumn{3}{c|}{CIFAR-10-LT} & \multicolumn{3}{c}{CIFAR-100-LT}  \\
\cmidrule{2-4} \cmidrule{5-7}
& \multicolumn{3}{c|}{Imb. Factor} & \multicolumn{3}{c}{Imb. Factor}\\
& 100 & 50 & 20 & 100 & 50 & 20 \\
\midrule
CE$^\dagger$                                                & 72.2 & 78.3 & 83.9 & 40.6 & 45.0 & 53.0 \\
Yu et al. (2022)  & 75.0 & 80.2 & 85.5 & 42.3 & 48.0 & 54.5 \\
DBM-CE                                                      & 74.4 & 80.0 & 85.5 & 41.1 & 46.5 & 54.7\\
DBM-BS                                                      & \B{79.3} & \B{83.8} & \B{87.0} & \B{45.0} & \B{50.0} & \B{56.6} \\
\bottomrule
\end{tabularx}
\caption{Top-1 accuracy (\%) of ResNet-32 for a comparative experiment to Yu et al. (2022). $^\dagger$ denotes results borrowed from Yu et al. (2022).}
\label{table:inst_difficulty}
\setlength{\tabcolsep}{1mm}
\end{table}

\begin{table}[t]
\centering
\begin{tabularx}{0.48\textwidth}{l|YYY}
\toprule
\multirow{2}{*}{Method} & \multicolumn{3}{c}{Imbalance Factor} \\
& 100 & 50 & 10 \\
\midrule
GLMC \cite{glmc_fei}    & 55.88 & 61.08 & 70.74 \\
DBM-BS & 55.17 & 60.11 & 70.97 \\ 
DBM-BCL & \B{56.17} & \B{61.36} & \B{71.28} \\ 
\bottomrule
\end{tabularx}
\caption{Top-1 accuracy (\%) of the ResNet-32x4d on CIFAR-100-LT with the imbalance factor of 100, 50 and 10.}
\label{table:cifar100_glmc}
\setlength{\tabcolsep}{1mm}
\end{table}

\paragraph*{Comparison on Wider ResNet.}
Global and local mixture consistency cumulative learning (GLMC) \cite{glmc_fei} is a recent approach that has demonstrated competitive performance in LTR by ensuring consistency between features obtained from mixed image via CutMix \cite{cutmix_yun} and MixUp \cite{mixup_zhang}.
However, the authors of GLMC used ResNet-32x4d architecture (an architecture with 4$\times$ inplanes) for their experiment on Long-tailed CIFAR.
Using the same architecture, DBM-BCL outperforms GLMC.
The comparison between GLMC, DBM-BS, and DBM-BCL on ResNet-32x4d on CIFAR-100-LT is presented in \cref{table:cifar100_glmc}

\subsection*{ImageNet-LT}

\begin{table}[h]
\centering
\begin{tabularx}{\linewidth}{p{0.4\linewidth}|YYY|Y}
\toprule
Method & Many & Med. & Few & All \\
\midrule
{\small LADE$^\dagger$ \cite{lade_hong}} & 65.1 & 48.9 & 33.4 & 53.0 \\
{\small BS$^\dagger$ \cite{bs_ren}}      & 65.8 & 53.2 & 34.1 & 55.4 \\
{\small PaCo$^\dagger$ \cite{paco_cui}}  & 64.4 & \B{55.7} & 33.7 & 56.0 \\
{\small BCL \cite{bcl_zhu}}              & 67.9 & 54.2 & 36.6 & 57.1 \\
\midrule
{\small DBM-BCL}                         & \B{68.3} & 54.3 & \B{38.9} & \B{57.6} \\
\bottomrule
\end{tabularx}
\caption{Top-1 accuracy (\%) of ResNeXt-50 trained for 180 epochs on ImageNet-LT. $^\dagger$ denotes results borrowed from \citeauthor{bcl_zhu} \shortcite{bcl_zhu}.}
\label{table:imgnet_180epoch}
\setlength{\tabcolsep}{1mm}
\end{table}

\paragraph*{Results from Extended Epochs.}
Following the baseline approach of BCL \cite{bcl_zhu}, we conducted experiments on ResNeXt-50 trained for 180 epochs on ImageNet-LT \cite{imagenetLT_liu} to compare performance with previous methods.
The results are shown in \cref{table:imgnet_180epoch}.
DBM-BCL achieves superior performance compared to all prior methods in terms of overall accuracy, surpassing BCL by 0.5\%p.
Notably, our method demonstrates performance enhancement across all groups without sacrificing accuracy in the `Many' group to improve accuracy in the `Few' group.

\begin{table}[t]
\centering
\begin{tabularx}{\linewidth}{p{0.5\linewidth}|YYY|Y}
\toprule
Method & Many & Med. & Few & All \\
\midrule
{\small CE$^\dagger$} & 64.0 & 33.8 & 5.8 & 41.6 \\
{\small $\tau$-norm \cite{decoupling_kang}} & 56.6 & 44.2 & 27.4 & 46.7 \\
{\small cRT \cite{decoupling_kang}} & 58.8 & 44.0 & 26.1 & 47.3 \\
{\small LWS \cite{decoupling_kang}} & 57.1 & 45.2 & 29.3 & 47.7 \\
{\small LDAM-DRW$^\ddagger$ \cite{ldam_cao}} & 60.4 & 46.9 & 30.7 & 49.8 \\
{\small CE-DRW$^\ddagger$ \cite{ldam_cao}} & 61.7 & 47.3 & 28.8 & 50.1 \\
{\small BS$^\ddagger$ \cite{bs_ren}} & 60.9 & 48.8 & 32.1 & 51.0 \\
{\small KCL \cite{kcl_kang}} & 61.8 & 49.4 & 30.9 & 51.5 \\
{\small TSC \cite{tsc_li}} & 63.5 & 49.7 & 30.4 & 52.4 \\
{\small DisAlign \cite{disalign_zhang}} & 61.3 & 52.2 & 31.4 & 52.9 \\
{\small RIDE \cite{ride_wang}} & 66.2 & 51.7 & 34.9 & 54.9 \\
{\small BCL$^\star$ \cite{bcl_zhu}} & 66.0 & 53.7 & 36.7 & 56.1 \\
{\small GML$^\star$ \cite{gml_suh}} & \B{66.7} & 54.7 & 36.8 & 57.2 \\
\midrule
{\small DBM-BCL} & 64.4 & 53.9 & 41.8 & 56.3 \\
{\small DBM-GML} & 65.3 & \B{55.1} & \B{43.1} & \B{57.4} \\
\bottomrule
\end{tabularx}
\caption{Top-1 accuracy (\%) of ResNet-50 on ImageNet-LT. $^\dagger$ and $^\ddagger$ denotes results borrowed from \citeauthor{decoupling_kang} \shortcite{decoupling_kang} and \citeauthor{cmo_park} \shortcite{cmo_park}, respectively. $^\star$ denotes reproduced results with the official code.}
\label{table:imgnet2}
\setlength{\tabcolsep}{1mm}
\end{table}

\paragraph*{Group-wise Accuracy on ImageNet-LT.}
In this section, we present group-wise accuracy to analyze in detail the performance improvement of applying the proposed technique to large datasets.
All techniques listed in \cref{table:imgnet2} were implemented on the ResNet-50 model and evaluated on ImageNet-LT.
DBM-BCL and DBM-GML outperform the baseline, demonstrating the highest performance compared to other techniques.
Group-wise analysis reveals that significant improvements within the `Few' group contribute to the overall performance gains.

\subsection*{DBM with LLM-based methods}
Recently, classification models incorporated with large language models (LLMs) have demonstrated state-of-the-art performances in long-tailed recognition, with superior representation ability achieved from visual-text pre-training \cite{lift_shi, lpt_dong}.
LIFT \cite{lift_shi} finetunes a foundation model like CLIP \cite{clip_radford} using the logit adjustment (BS) objective, which makes it compatible with our method. \Cref{tab:lift} shows the experimental results of our method applied with on CIFAR-100-LT (IF 100) and ImageNet-LT using a ViT-B/16 encoder \cite{vit_dosovitskiy}.
Although the overall improvement appears marginal, our method effectively mitigates class imbalance, demonstrating significant gains in the `Few' group.

\begin{table}[h]
\centering
\setlength{\tabcolsep}{1mm}
\begin{tabularx}{\linewidth}{l|YYY|Y}
\toprule
Method & Many & Medium & Few & Overall \\
\midrule
\multicolumn{3}{l}{\textit{ImageNet-LT}} \\
\midrule
BS      & 80.2 & 75.9 & 71.3 & 76.9\\
DBM-BS  & 78.9 & 75.9 & 75.1 & 77.0\\
\midrule
\multicolumn{3}{l}{\textit{CIFAR-100-LT (Imb. Factor=100)}} \\
\midrule
BS   & 84.0 & 80.5 & 73.8 & 79.7\\
DBM-BS & 83.5 & 80.0 & 77.0 & 80.3\\
\bottomrule
\end{tabularx}
\caption{Top-1 accuracy (\%) of finetuned CLIP (ViT-B/16 encoder) on ImageNet-LT and CIFAR-100-LT with imbalance factor of 100.}
\label{tab:lift}
\setlength{\tabcolsep}{1mm}
\end{table}

\end{document}